\documentclass[sigconf]{acmart}

\usepackage{booktabs} % For formal tables
\usepackage{enumitem}
\usepackage[ruled]{algorithm2e}
\usepackage{url}

\newenvironment{myquote}%
  {\list{}{\leftmargin=0.1in\rightmargin=0.3in}\item[]}%
  {\endlist}

\setcopyright{rightsretained}

\copyrightyear{2020}
\acmYear{2020}
\setcopyright{licensedothergov}\acmConference[GECCO '20]{Genetic and Evolutionary Computation Conference}{July 8--12, 2020}{Cancún, Mexico}
\acmBooktitle{Genetic and Evolutionary Computation Conference (GECCO '20), July 8--12, 2020, Cancún, Mexico}
\acmPrice{15.00}
\acmDOI{10.1145/3377930.3389840}
\acmISBN{978-1-4503-7128-5/20/07}

\begin{document}

\title{Novelty Search makes Evolvability Inevitable}

\author{Stephane Doncieux}
\affiliation{\institution{Sorbonne Universit\'{e}, CNRS, Institut des Syst\`{e}mes Intelligents et de Robotique, ISIR, F-75005 Paris, France}}
\email{stephane.doncieux@sorbonne-universite.fr}

\author{Giuseppe Paolo}
\affiliation{\institution{AI Lab, SoftBank Robotics Europe, Paris, France} \institution{Sorbonne Universit\'{e}, CNRS, Institut des Syst\`{e}mes Intelligents et de Robotique, ISIR, F-75005 Paris, France}}
\email{giuseppe.paolo@softbankrobotics.com}

\author{Alban Laflaqui\`{e}re}
\affiliation{\institution{AI Lab, SoftBank Robotics Europe, Paris, France}}
\email{alaflaquiere@softbankrobotics.com}

\author{Alexandre Coninx}
\affiliation{\institution{Sorbonne Universit\'{e}, CNRS, Institut des Syst\`{e}mes Intelligents et de Robotique, ISIR, F-75005 Paris, France}}
\email{alexandre.coninx@sorbonne-universite.fr}

\begin{abstract}
Evolvability is an important feature that impacts the ability of evolutionary processes to find interesting novel solutions and to deal with changing conditions of the problem to solve.
The estimation of evolvability is not straight-forward and is generally too expensive to be directly used as selective pressure in the evolutionary process.
Indirectly promoting evolvability as a side effect of other easier and faster to compute selection pressures would thus be advantageous.
In an unbounded behavior space, it has already been shown that evolvable individuals naturally appear and tend to be selected as they are more likely to invade empty behavior niches.
Evolvability is thus a natural byproduct of the search in this context.
However, practical agents and environments often impose limits on the reachable behavior space. How do these boundaries impact evolvability? In this context, can evolvability still be promoted without explicitly rewarding it?
We show that Novelty Search implicitly creates a pressure for high evolvability \textit{even in bounded behavior spaces}, and explore the reasons for such a behavior.
More precisely we show that, throughout the search, the dynamic evaluation of novelty rewards individuals which are very mobile in the behavior space, which in turn promotes evolvability.

\end{abstract}




\maketitle

\section{Introduction}
Evolutionary Robotics studies the automatic design of robotic agents through evolutionary mechanisms \cite{bongard2013evolutionary, doncieux2015evolutionary}.
Exploration has been shown to be a critical issue in this field \cite{lehman2008exploiting, mouret2009overcoming, lehman2011abandoning}.
The search process at its core needs to be able to generate diverse individuals, among which the most relevant can be selected according to their performance or to promote exploration \cite{doncieux2014beyond}.
The ability of the search process to generate creative individuals can be captured by the notion of \textit{evolvability}.
Different definitions of evolvability have been proposed in the literature to characterize the potential creativity of the evolutionary process.
This potential can be defined by the ability of the process to generate individuals that are diverse \cite{kirschner1998evolvability,lehman2011improving} and possibly efficient \cite{altenberg1994evolution, tarapore2015evolvability, tarapore2016different}, eventually as an answer to an adaptive challenge \cite{tarapore2015evolvability}.

Estimating the evolvability of an individual is not straightforward.
Some algorithms estimate it via sampling \cite{mengistu2016evolvability}, which requires a huge amount of costly evaluations.
Finding a selective pressure that would be simple and cheap to compute while indirectly fostering evolvability is thus of critical interest.
Several properties or mechanisms have already been shown to increase evolvability: a pressure on neural network connection cost \cite{clune2013evolutionary}, fitness landscape ruggedness \cite{kounios2016resolving}, extinction events \cite{lehman2015extinction}, and divergent selection \cite{lehman2016critical}.
Among these different approaches, divergent selection is of particular interest as its general formalization imposes little constraints on the phenotype, problem dynamic or ruggedness.
Any breakthrough on this approach would thus apply to a large number of contexts.

Novelty Search (NS) is one of the main divergent search algorithms \cite{lehman2011abandoning}.
Its ability to promote evolvability has already been observed in different contexts \cite{lehman2011improving,wilder2015reconciling}, but the reasons why selecting for novelty also fosters evolvability are still unclear.
In the case of an unbounded environment, it has been shown that evolvability is a natural byproduct of evolution, as evolvable individuals tend to invade more niches than their competitors \cite{lehman2013evolvability}.
However, when considering realistic robotic applications, the behavior space reachable by the physically limited robot is generally bounded.
Still, it appears that NS successfully generates a high level of evolvability even in such bounded behavior spaces \cite{lehman2011improving}.
The behavior is not surprising at the beginning of the search, when evolvable individuals can spread due to the "niche founder" effect \cite{lehman2013evolvability}, but is not fully explained once the boundaries of the behavior space have been reached.
We hypothesize that, even after that point, NS continues pushing towards evolvable individuals.
It is not trivially apparent that such behavior would occur in practice, as NS has been shown to tend towards a uniform sampling of the behavior space \cite{doncieux2019novelty}. Once all areas of the behavior space have been reached, what is the dynamic? Is there still a pressure towards evolvability? If so, why?

In this work, we consider the \textit{evolvability} of an individual (or set of individuals) as the combination of two features relative to the new individuals it can generate through mutations:
(1) how much of the reachable behavior space they cover and (2) how uniformly they do so.
We experimentally show that NS generates evolvable individuals, even long after the boundaries of the reachable behavior space have been reached.
We hypothesize that this behavior is due to the dynamic nature of the novelty criterion, and investigate the following explanations of the phenomenon:
\begin{enumerate}
    \item NS pushes individuals to constantly move in the behavior space: in new and unexplored areas first, \textit{but also then in already explored areas} as their density of individuals is never exactly homogeneous, 
    \item the dynamic nature of the reference set used to compute novelty (population and archive) has a causal role on keeping evolvability high.
\end{enumerate}

\section{Novelty Search}

In the following, we will use the notations introduced in \cite{doncieux2019novelty}. The main feature of NS is to replace the usual goal-oriented objective driving the evolutionary process by a criterion measuring \textit{novelty} as the average distance of an individual to its closest neighbors in a behavior space:
\begin{equation}
 \rho(x) = \frac{1}{k} \sum_{i=1}^k \text{dist} \big( \varphi_{\mathcal{B}}(x), \varphi_{\mathcal{B}}(v_i) \big),
 \label{eq:novelty}
\end{equation}
where $x$ is the \textit{genotype}, i.e. an individual, the $v_i$ are its $n$ closest neighbors among an archive of previously explored individuals and the current population in a behavior space $\mathcal{B}$, and $\phi_\mathcal{B}(.)$ is the function mapping a genotype to its associated behavior descriptor\footnote{It should be noted that, besides the genotype, this function depends on many parameters of the evaluation, like the initial state of the robot, the evaluation length and also, of course, the environment itself. All these elements will be considered as constant here.}.

Beyond this objective, NS does not impose strong constraints on the evolutionary algorithm used to generate genomes.
The first experiments implementing NS relied on NEAT \cite{lehman2011abandoning}, but other works have used evolutionary algorithms like NSGA-II --which has the advantage of not being specific to neuroevolution-- together with a simpler encoding \cite{mouret2012encouraging}.
Likewise, several strategies have been used to manage the archive of previously explored individuals, either adding the most novel individuals \cite{lehman2011abandoning} or adding randomly chosen ones \cite{gomes2015devising}. 

In the following, we have chosen to use the simplest algorithm possible. We have considered a neural network with a fixed structure and a simple elitist selection algorithm that generates $\lambda$ individuals at each generation and keeps the $\mu$ most novel ones.

\begin{algorithm}
\SetAlgoLined
\textbf{Inputs:} population size $\mu$, number of offspring $\lambda$, environment \textbf{env}, number of neighbors for novelty calculation $k$, number of generations $G$\;
\KwResult{Generated policies.}
\textbf{pop} $\leftarrow$ \textit{RandomPopulation($\mu$)}\;
\textbf{arch} $\leftarrow \emptyset$\;
\textbf{gen} = 0\;
\While{\textbf{gen} < $G$}{
    \textbf{off} $\leftarrow$ \textit{generateOffspring}(\textbf{pop}, $\lambda$)\;
    \For{\textbf{agent} in (\textbf{pop} $\cup$ \textbf{off})}{
        (\textbf{agent.fit}, \textbf{agent.bd})=\textit{evaluate}(\textbf{agent},\textbf{env})
    }
    \textbf{novRefSet} $\leftarrow$ \textbf{pop} $\cup$ \textbf{off} $\cup$ \textbf{arch}\;
    \For{\textbf{agent} in (\textbf{pop} $\cup$ \textbf{off})}{
        \textbf{agent.novelty} $\leftarrow$ \textit{getNovelty}(\textbf{agent.bd}, \textbf{novRefSet}, $k$)\;
    }
    \tcc{Update archive, either with random samples or with the most novel ones:}
    \textbf{arch} $\leftarrow$ \textbf{arch} $\cup$ \textit{sample}(\textbf{off})\;
    \textbf{pop} $\leftarrow$ \textit{selectMostNovel}(\textbf{pop} $\cup$ \textbf{off}, $\mu$)\;
    \textbf{gen} $=$ \textbf{gen} $+ 1$\; 
}

 \caption{Novelty search}
 \label{alg:ns}
\end{algorithm}

\section{Evolvability: Definition and estimation}
\label{sec:evolv}
Evolvability is a central topic in evolutionary processes, may it be in biology \cite{pigliucci2008evolvability} or in evolutionary computation \cite{hu2010evolvability}.
It has consequently attracted a lot of attention and has been given different definitions. For Hu and Banzhaf, it is \textit{"the capability of a system to generate adaptive phenotypic variation and to transmit it via an evolutionary process"} \cite{hu2010evolvability}.
This view actually highlights the two main facets of evolvability, which are (1) to generate a variability that (2) may lead to adaptation.
The ability to adapt to changes in the fitness landscape is a measure of what is actually expected from evolvability, and it has been the focus of many works \cite{reisinger2005towards,clune2013evolutionary,shorten2014evolvable, tarapore2015evolvability}. 

In evolutionary robotics, the adaptation ability can be empirically tested by changing the evaluation conditions of the robot, may it be its morphology, including its motors \cite{tarapore2015evolvability}, or its environment \cite{shorten2014evolvable}.
From a theoretical perspective, selecting the conditions to which to adapt to is actually not straightforward. Too simple environmental changes may not need any adaptation, while too difficult ones may be out of reach \cite{shorten2014evolvable}.
To avoid this issue, in the following we will focus on the ability to generate diverse behaviors, considering that, if an adaptive challenge appears, a system with a greater ability to generate variation is more likely to successfully adapt.

For this assumption to hold, the space in which the variations are observed needs to be carefully chosen.
This is the motivation for recent approaches which propose to automatically build it \cite{cully2019autonomous, paolo2019unsupervised}. Regardless of its origin, this space needs to make sense with respect to the kind of task the robot could have to achieve.
If it does not capture at all the variations that could lead to an adaptation, this assumption clearly does not hold.
As a consequence, we will consider variations in a behavior space that is \textit{aligned} with what is expected from the robot \cite{pugh2016quality}.
For example, in maze navigation experiments, the end position of the robot in the maze is an indirect indication of the navigation capability of the robot. On the contrary, the final orientation of the robot is not aligned with the navigation capability, as a robot simply turning round without changing position could reach any behavior characterized this way without exploring the maze.
To highlight the importance of this characterization, we will call such a behavior space an \textit{outcome space}.

To adapt to new conditions that are not known in advance, the reachable part of the outcome space needs to be as large as possible, to maximize the chances of rapidly discovering solutions to the new challenge, or at least stepping stones leading to such solutions.
Another important aspect is that each possible behavior needs to have similar chances to be reached. In the absence of any clue about what part of the outcome space is likely to be useful to deal with the new adaptive challenge, the best strategy is to generate a sampling of the reachable space that is as uniform as possible, so that, on average, the number of generations required to discover a solution, no matter where it is, is as low as possible.
It leads to the definition of two different criteria to estimate evolvability: (1) a measure of the reachability and (2) a measure of the sampling uniformity.
\\

Evolvability is an estimation of the \textit{potential to generate variability}. Consequently, we estimate it on the basis of a large sampling of offspring: for each individual, a significant number of offspring is generated by applying the evolutionary process mutation operator. 
Those sets are then used to produce estimations of \emph{reachability} and of \emph{uniformity}, as detailed below, both per individual and at a population level \cite{lehman2016critical}. For individual metrics, we analyze the set of generated offspring of each individual in the population individually. For population level metrics, we merge those sets together and perform computations on the superset. Individual metrics allow us to explore the ability of an algorithm to produce highly evolvable individuals, whereas population level metrics show the adaptation capability of the whole evolutionary process. It measures the complementarity of individuals and estimates how well the algorithm pushes individuals to spread over the reachable space.

The reachability corresponds to the portion of the outcome space that can be reached by the offspring.
Such a measure is inspired from MAP-Elites \cite{mouret2015illuminating} and has frequently been used to characterize the exploration ability of divergent search algorithms \cite{cully2017quality}.
In practice, it is computed by dividing the outcome space into regular grid cells and by counting the ratio of those cells that are reached by at least one offspring:
$$n_r(\mathcal{O}) =\big| \{c_i > 0 \} \big|,$$
$$R(\mathcal{O}) = \frac{n_r(\mathcal{O})}{n},$$
where $\mathcal{O}$ is the set of offspring, $|.|$ denotes the cardinality,  $c_i$ is the number of offspring in cell $i$, $n$ is the total number of cells, and $n_r(\mathcal{O})$ is the number of cells reached by at least one offspring.
The uniformity measure is based on the same grid cells and is derived from the Jensen-Shannon Distance \cite{endres2003new} between the discrete distribution of the offspring in the grid and its theoretical uniform counterpart, as done in Gomes et al. \cite{gomes2015devising}:
$$U(\mathcal{O}) = 1 - JSD(P_\mathcal{O}, Q),$$
where $P(\mathcal{O})$ is the discrete distribution of the offspring over the cells:
$$ P(\mathcal{O}) = \left(\frac{c_1}{|\mathcal{O}|}, \dots, \frac{c_i}{|\mathcal{O}|}, \dots \frac{c_{n_r(\mathcal{O})}}{|\mathcal{O}|}\right),$$
$Q$ is the uniform distribution over the same cells:
$$ Q=\left( \frac{1}{n_r(\mathcal{O})}, \dots, \frac{1}{n_r(\mathcal{O})} \right)$$
and $JSD(P_\mathcal{O}, Q)$ is defined as:
$$
JSD(P_\mathcal{O}, Q)=\sqrt{\frac{D_{KL}\left(P_\mathcal{O} \middle|\middle| M \right) + D_{KL}\left(Q \middle|\middle| M\right)}{2}},
$$
with $M = \frac{P_\mathcal{O}+Q}{2}$ and $D_{KL}(\cdot || \cdot)$ being the Kullback-Leibler divergence.

Contrary to Gomes et al. \cite{gomes2015devising}, only cells with a non-null number of offspring are taken into account to define $P$ and $Q$. This choice has two motivations: the measure can easily be extended to unbounded behavior spaces, and it prevents reachability and uniformity measures to overlap in terms of information content.
Reachability measures how much of the outcome space is covered by the offspring, while uniformity measures to what extent this coverage is uniform. If all cells are taken into account, the Jensen-Shannon Distance also estimates the coverage. While this measure alone could be sufficient, we think that separating the two aspects makes interpretation easier as a same Jensen-Shannon Distance can be obtained with set of points having a different coverage.

Note that regardless of the individual or population level of analysis, the extra offspring generated this way are only used for evolvability measures and are ignored by the evolutionary process that drives the search.

\begin{figure}
    \centering
    \includegraphics[width=\linewidth]{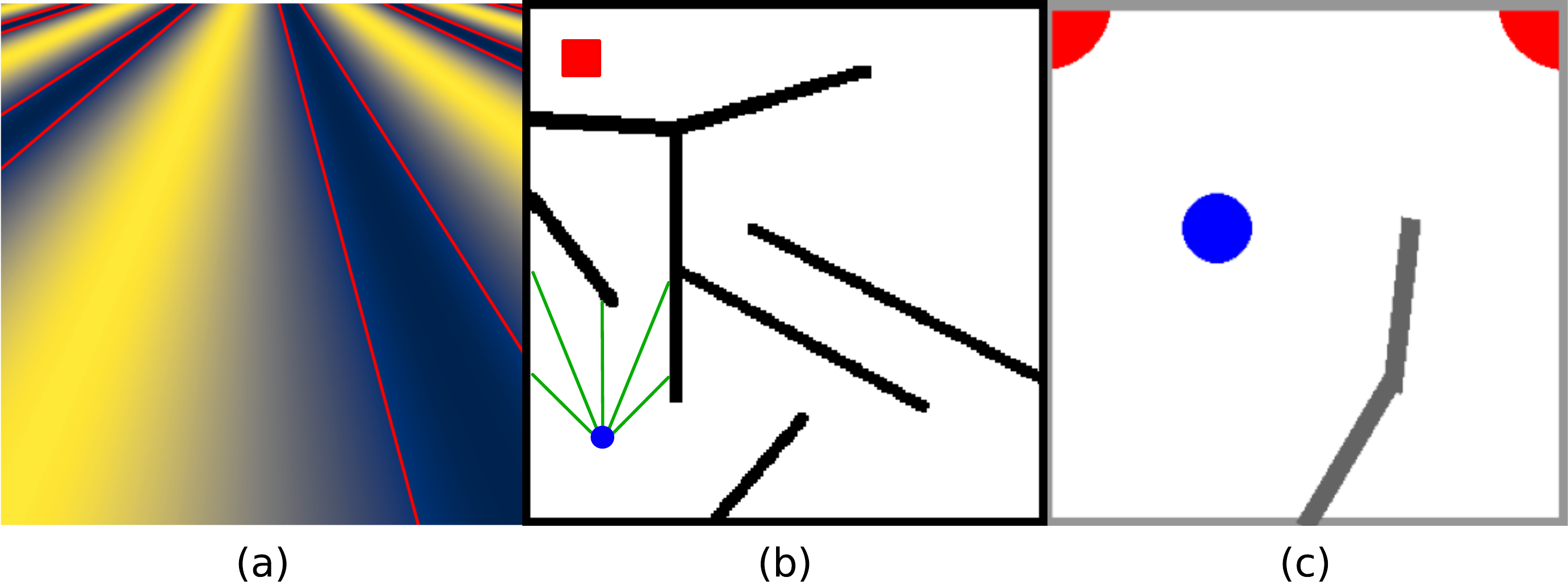}
    \caption{Environments. (a) A toy problem in which the outcome space is a sinusoidal function of the genome. (b) The hard maze introduced in \cite{lehman2008exploiting}. (c) A 2-jointed robotic arm pushing a ball.}
    \label{fig:envs}
\end{figure}

\section{Experimental setup}

In order to formulate and test hypotheses about the impact of NS on evolvability, we introduce different experimental setups and variants of NS algorithms.
The detailed rationale for the choice of those variants and the analysis of the experimental results will be presented in section \ref{sec:NS and evolvability}.

\subsection{Simulations}

Three simulated environments were considered:
\begin{itemize}[leftmargin=*]
\item \emph{Toy Problem}:
a simple synthetic environment allowing a fast and straightforward analysis of the search. It consists of a redundant sinusoidal mapping from a 2D genome space to a 1D outcome space, such that its phase and period are controlled by the 2 dimensions of the genome space respectively:
$$ \varphi_{\mathcal{B}}(x_1, x_2) = - 5 \cdot \sin \left(\frac{2 \pi x_1}{11 - 1.8 x_2} \right).$$
The variable periodicity of the mapping allows different genomes to map to the same behavior, while having different evolvabilities. The value of the behavior associated with each genome is represented in the genome space in figure \ref{fig:envs}.(a);
\item \emph{Hard maze}:
an environment introduced in \cite{lehman2008exploiting} in which an agent explores a maze, as illustrated in \ref{fig:envs}.(b).
The two-wheeled agent is equipped with $5$ distance sensors in the front, shown in green, and two bumpers in the front.
The policy controlling the speed of each wheel is parametrized by a $2$-layers fully connected neural network, with 10 neurons per layer, and that takes as input the sensors readings. The genome consists of the neural network weights, and the outcome space is the final $(x, y)$ position of the robot after 2000 timesteps;
\item \emph{Robotic arm}:
an environment in which a 2-jointed robotic arm can push a ball in a square arena, as shown in figure \ref{fig:envs}.(c). The arm is controlled by a $2$-layers fully connected neural network, with 10 neurons per layer, taking as input the state of the environment (ball position, joints angles and velocities).
The genome consists of the neural network weights, with genome space being identical to the hard maze experiment.
The outcome space consists of the final $(x,y)$ position of the ball after $300$ timesteps.
\end{itemize}
Although NS does not require a goal and fitness function in the outcome space, some were nonetheless defined in each environment in order to compare its behavior to fitness-based approaches in section \ref{sec:NS and evolvability}. The fitness function is defined as the Euclidean distance from the outcome to the goal in both the toy problem and hard maze experiments, and as a binary success value in the robotic arm experiment depending on if the ball lands in the goal or not.
The goals are represented in red in figure \ref{fig:envs}. 
Note that, for the toy problem the outcome space is 1D in the range $[-5, 5]$ and the goal corresponds to $-4$.

\begin{figure*}[thp]
    \includegraphics[width=\textwidth]{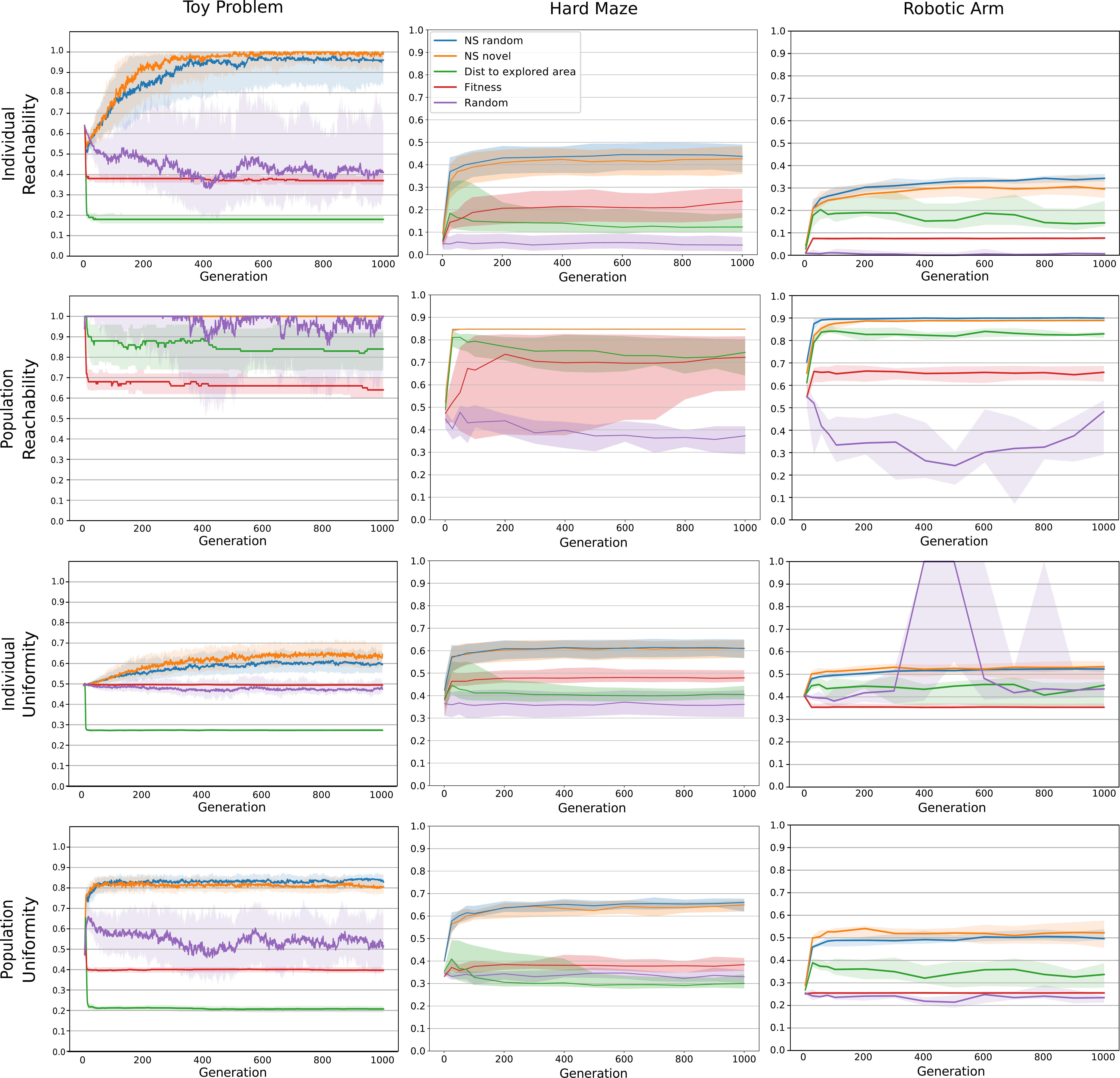}
    \caption{Reachability and uniformity metrics at the individual and population level for all of the search algorithm variants in the three experimental environments. Median, first and third quartiles are shown for 100 runs on the Toy Problem and 20 on the Hard Maze and on the Robotic Arm environments.}   
    \label{fig:plots_h1}
\end{figure*}

\subsection{Evolutionary algorithms variants}

Five variants of search algorithms were run in all environments:
\begin{itemize}[leftmargin=*]
\item \emph{NS novel}: the vanilla NS algorithm described in Algorithm \ref{alg:ns}, with the most novel offspring added to the archive of genomes;
\item \emph{NS random}: another NS algorithm where random offspring are added to the archive of genomes;
\item \emph{Distance to Explored Area (DEA)}: a divergent search algorithm in which novelty is replaced by the distance to the convex hull \cite{edelsbrunner1983shape} of individuals already explored in the outcome space (counted negatively if inside the hull and positively if outside), to be maximized;
\item \emph{Random}: a variant in which the parents of the next generation are randomly selected among the offspring and the current parents;
\item \emph{Fitness}: a fitness-based elitist algorithm in which novelty is replaced by the fitness defined in the outcome space.
\end{itemize}
The \emph{DEA} variant tends to make the reachable space as large as possible, but compared to the NS variants, it strongly discourages the search to go back inside the hull, including already visited areas.
The \emph{Random} variant does not take into account the novelty of the genomes during the search, and is only driven by random drift.
Finally, the \emph{Fitness} variant does not take novelty into account either but only the fitness of each individual.

Each evolutionary algorithm variant is run for $1000$ generations on each environment\footnote{Each variant is run multiple times, in order to assess the statistical significance of the results.}.
The population size is set to $\mu=20$ on the toy problem, due to its low-dimensional nature, while it is set to $\mu=100$ in the hard maze and robotic arm problems. At each generation, $\lambda=40$ offspring are generated for the toy problem, and $\lambda=200$ for the hard maze and for the robotic arm problems.
Similarly, the polynomial bounded mutation \cite{deb2002fast} operator is used in all runs with $\eta=45$ for the toy problem and $\eta=15$ for the hard maze and robotic arm problem. The minimum and maximum bounding values for each gene are set to $-5$ and $5$.
Finally, the number of neighbors considered to compute novelty in the NS variants is set to $k=10$ for the toy problem and to $k=15$ in the other two problems.

In order to evaluate reachability and uniformity (see Section~\ref{sec:evolv}), the outcome spaces of the environments are discretized with a resolution of $50$. The 1D outcome space of the toy problem is thus split into $50$ regular cells, while the 2D outcome spaces of the hard maze and robotic arm environments are split into $50 \times 50 = 2500$ cells.
The number of offspring $|\mathcal{O}|$ generated per genome for these measures is set to $1000$ for the toy problem and $5000$ for the other two, which corresponds to $2$ genomes per cell in the case of an ideal uniform distribution.

The source code of the experiments is available at \linebreak \url{https://github.com/alaflaquiere/simple-ns} for the toy experiment, and \url{https://github.com/robotsthatdream/diversity_algorithms} for the other two experiments.

Throughout this paper, in order to compare different conditions and evaluate the statistical significance of the results, the datasets were compared by performing pairwise two-tailed Mann-Whitney tests~\cite{mann1947test}.
In the case of multiple comparisons, the Holm-Bonferroni correction~\cite{holm1979simple} was applied for each figure on the tests of all the algorithm variants.

\section{NS \& evolvability}
\label{sec:NS and evolvability}

In this section, we analyze the relation between NS and evolvability, first by formalizing empirical observations, second by hypothesizing the dynamic behind these observations, and third by testing these hypotheses.  

\subsection{Novelty Search promotes evolvability}

Figure \ref{fig:plots_h1} shows the evolution of the reachability and uniformity, at an individual and population level, for all search variants and all environments.
It allows us to formulate the following observation:
\begin{myquote}
\emph{\textbf{Observation:} Novelty Search pushes towards high evolvability throughout the search.}
\end{myquote}
Indeed, as can be seen in figure \ref{fig:plots_h1}, reachability and uniformity at an individual and population level increase and remain high throughout the run for both \emph{NS novel} and \emph{NS random} 
(when comparing the two NS variants: 
\emph{individual reachability} on hard maze and robotic arm: $p\leq8.9 \cdot 10^{-4}$, not statistically significant on the toy problem; 
\emph{population reachability} not statistically significant on the three environments;
\emph{individual uniformity} on toy problem: $p\leq8.74 \cdot 10^{-3}$, not statistically significant on the hard maze and the robotic arm;
\emph{population uniformity} on toy problem: $p\leq6.17 \cdot 10^{-3}$, not statistically significant for hard maze and robotic arm).

The similar profiles of these two variants also indicate that the strategy chosen to add individuals to the archive has little impact on the evolution of evolvability.
We can also see that the population reachability quickly reaches its maximum value\footnote{Because of the configuration of the environment, some cells might not be reachable; in those cases, the maximum value is thus lower than 1 in practice.} and exhibit very little variability, which indicates that the genomes in the population quickly become complementary in covering all areas of the reachable outcome space, even when no individual can cover it on its own.
This phenomenon is confirmed by the high level of uniformity reached with respect to other variants (when comparing NS variants to the other variants: \emph{individual reachability}: $p\leq1.09 \cdot 10^{-20}$; \emph{population reachability} on hard maze and robotic arm: $p\leq6 \cdot 10^{-3}$, not statistically significant for the toy problem; \emph{individual uniformity}: $p\leq1.48 \cdot 10^{-6}$; \emph{population uniformity}: $p\leq1.07 \cdot 10^{-2}$).

The other search algorithm variants considered in our experiments show very different dynamics than NS.
In the case of \emph{DEA}, the reachability sometimes increases at the beginning of the run, but then tends to decrease significantly. This behavior is expected as the selection pressure for this variant pushes the genomes towards the borders of the outcome space. Once the borders have been reached, genomes with less evolvability are favored as they tend to stay on the borders of the outcome space.
In the case of the \emph{Fit} and \emph{Random} variants, reachability and uniformity fluctuate significantly, while remaining lower than for the NS variants. This indicates that these variants tend to focus their exploration on only a subpart of the outcome space, and that the high evolvability generated by NS is not due to an intrinsic property of the environment or of its fitness function, but is due to the search for novelty.

\begin{figure*}[htp]
    \includegraphics[width=\textwidth]{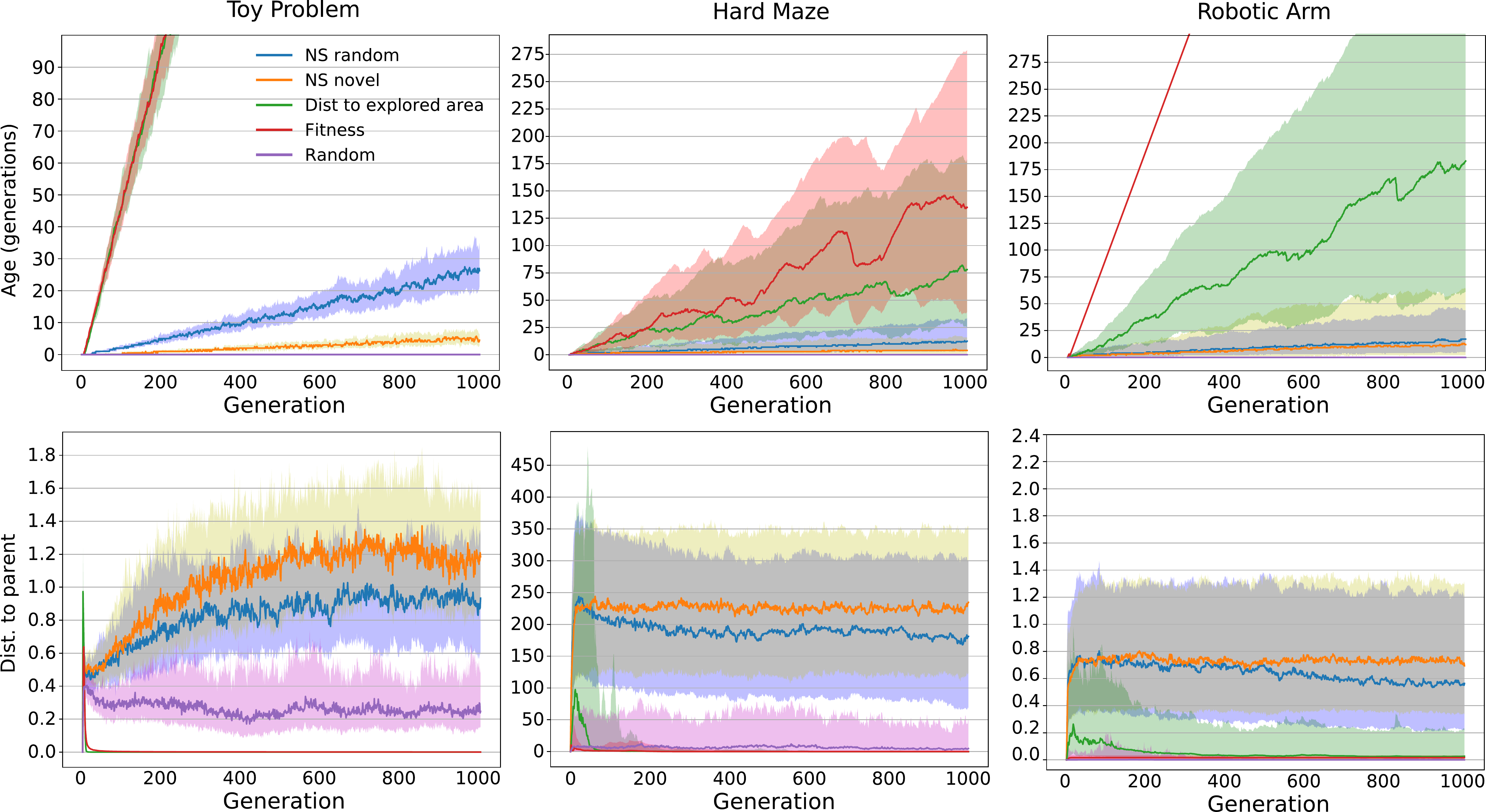}
    \caption{Age of individuals and distance to parent on the three experimental environments for all of the search algorithm variants. Median, first and third quartiles are shown for 100 runs on the Toy Problem, 20 on the Hard Maze and on the Robotic Arm environments.}   
    \label{fig:ages}
\end{figure*}

\subsection{Evolvability results from the dynamic of the novelty landscape}

Given these empirical observations, we now formulate and test hypotheses to explain how NS promotes evolvability.

\subsubsection{NS promotes a dynamic population}

Novelty Search has the particular property of selecting genomes based on a dynamic measure: novelty. Once the boundaries of an outcome space have been reached, NS thus drives the population to re-explore previously explored areas in which novelty can appear higher than in more recently explored ones.
As a consequence, we formulate the following hypothesis:
\begin{myquote}
\emph{\textbf{Hypothesis 1:} NS maintains a dynamic and ever moving population, where selected individuals are highly different from one generation to the next in the outcome space.}
\end{myquote} 
If true, this property should in turn favor the selection of genomes with higher evolvability throughout the search, as their ability to move in the outcome space is higher.

In order to validate this hypothesis, we add two new metrics to characterize the individuals in a population: their age, which is the difference between the current generation and the generation at which they were generated, and the Euclidean distance to their parent in the outcome space. A dynamic population should be renewed often (low ages), and with children that are behaviorally different from their parent (high distance to parent).

The results, displayed in figure~\ref{fig:ages}, indeed show that for both NS variants, the distance to parent is significantly higher that for other methods, and remains high even after the outcome space has been fully explored, while the age remains significantly lower 
(when comparing the two NS variants: \emph{distance to parent}: $p\leq3.14 \cdot 10^{-3}$; 
\emph{age} on the toy problem and the hard maze: $p\leq1.5 \cdot 10^{-33}$, not statistically significant on the robotic arm.
When comparing the NS variants to the other variants:
\emph{distance to parent}: $p\leq3.74 \cdot 10^{-18}$; 
\emph{age}: $p\leq2.55 \cdot 10^{-33}$). 

The plots highlight a difference between NS variants: \emph{NS random} tends to have older individuals and with a smaller distance to their parents than \emph{NS novel}. This is expected as the most novel individuals have a higher chance to stay novel for a while as they won't be systematically added to the novelty reference set. In \emph{NS novel}, they are added to the archive and thus can't stay novel for long and have to move more rapidly.

By contrast, with the other search algorithms, the distance to parent is high during the initial exploration phase, but it plummets once the algorithm has maximized its objective, i.e. has found the goal for \emph{Fitness} or has reached the borders of the closed environment with \emph{DEA}, while the age skyrockets as there is no pressure to renew the already high performing individuals in the population (the selection scheme is elitist). 

The random selection baseline also shows significantly lower distances to parent than the NS variants despite renewing the population very often, which demonstrates that population dynamism is specifically promoted by NS and not a sole consequence of a random drift. 

\begin{figure*}[thp]
    \includegraphics[width=0.67\textwidth]{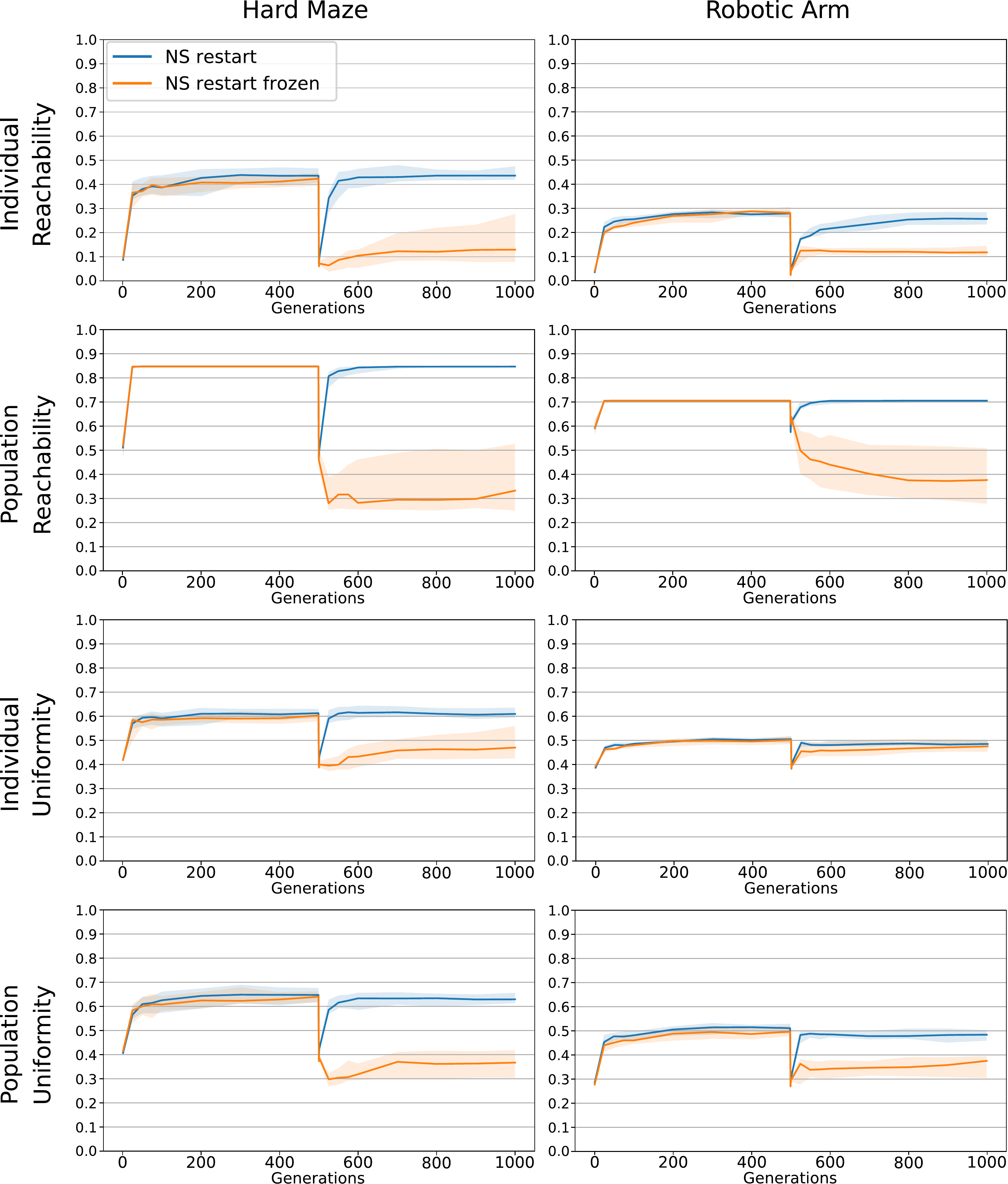}
    \caption{\emph{NS Random} with restart of the population vs \emph{NS Random} with restart of the population and a frozen reference set. The plots show the reachability and uniformity metrics on the individual and the population level in the three experimental environments. Median, first and third quartiles are shown for 20 runs on the Hard Maze and the Robotic Arm environments.}   
    \label{fig:restart}
\end{figure*}

\subsubsection{Impact of the reference set}

We showed that NS makes the population very dynamic in the outcome space, and we gave the intuitive explanation that NS tends to re-explore areas once their novelty gets high again with respect to the rest of the outcome space.
For this phenomenon to happen, it is important that the reference set used to compute novelty be regularly updated. When it is, the recently explored areas get over-represented in the reference set and their novelty locally decreases; which in turns pushes the population to move away.
Having a dynamic reference set should thus be essential to promote evolvability.

\begin{myquote}
\emph{\textbf{Hypothesis 2:} The dynamism of the reference set of individuals from which the novelty of new individuals is computed is essential to promote evolvability.
}
\end{myquote}

To validate this hypothesis, we ran extra experiments with the \emph{NS Random} algorithm to explore its behavior if the reference set used to compute the novelty does not change. We do not include results on the toy problem, for which even the \emph{Random} variant reaches the maximum population reachability, thus showing that the selection pressure may not be critical in this environment, probably because of the small outcome space size (1D). 

To explore the impact of the reference set, the algorithm is first run normally for a fixed number of generations, in order to create a reference set (archive and population). This reference set is then saved, and the population is reinitialized with randomly generated individuals. The NS algorithm resumes from this point, but using the saved reference set to compute the novelty scores in the subsequent generations: the archive is never updated, and the current population is not used for novelty computation. The reference set is therefore populated, but remains static after the reinitialization. As a control condition, we run the same experiment, including the population reinitialization, but without freezing the reference set.

The results shown in figure~\ref{fig:restart} validate our hypothesis and showcase the importance of a dynamic reference set for NS. In both conditions, reachability and uniformity metrics experience a sharp drop after the reinitialization, as the previous population is replaced with randomly generated individuals that are not especially evolvable. In the control condition, those metrics quickly recover as the reference set follows the new population and selects evolvable individuals. In the frozen reference set condition, however, reachability and uniformity remain low, as the algorithm simply selects individuals that are considered novel with regard to the frozen reference set.

\section{Discussion}

The fact that NS can incentivize the re-exploration of already explored areas of the outcome space is important to maintain evolvability, as \emph{DEA}, that does not have such an incentive, is unable to maintain high evolvability throughout the run. Fitness does not warrant high evolvability, giving both lower coverage and uniformity than the NS variants. This is due to the fact that once areas with high fitness have been reached, the evolution process has no incentive in maintaining high evolvability for the remainder of the run. Selecting the genomes independently from their position in the outcome space, as in the \emph{Random} variant case, produces individual with low evolvability that tend to move only in a small and local area of the outcome space. This highlights the importance of using a proper selective pressure measure to foster evolvability. We have also highlighted the dynamic nature of the novelty criterion that pushes individuals to move in the outcome space. One of the key ingredient of this dynamic is the reference set used to compute novelty. As it contains both an archive and the current population, it changes at every generation as the archive grows and as the population changes. We have shown that this ever moving fitness landscape is responsible for the pressure towards evolvability. Besides that, the two archive growth strategies tested (random or most novel individuals) lead to a similar reach of the whole outcome space.

Other works have reached opposite conclusions and have shown that NS may result in a poor evolvability, at least lower than for experiments driven by the fitness \cite{shorten2014evolvable,lehman2011improving}. This seems to be a strong contradiction, but we actually think that it is compatible with the model of NS we have proposed. Our experiments show that evolvability in NS results from a strong push towards movement in the outcome space. Even long after the boundaries of the reachable space have been reached, individuals in the population are, and remain, young: the oldest ones are not kept and the new individuals that are selected are always far from their parents in the outcome space.  Individuals that survive are those that can reach an area $a$ considered as novel at a given moment while its parent was equally novel, but in another area that is likely far from $a$. It is this long traveled distance in a short number of generations that likely creates the pressure towards evolvability. This is the \textit{strategy} used by NS to push towards evolvability: by forcing individuals to constantly move, the most evolvable ones are indirectly favored by evolution.

In domains for which solutions are too fragile, as in the biped locomotion \cite{lehman2011improving} or in maze navigation with narrow corridors \cite{shorten2014evolvable}, this strategy may just fail. If generated individuals have trouble to "go" from one novel area to another, the dynamic we have observed will not work. One factor may be the number of generated samples. For the maze experiment, the population size was relatively small (25 individuals) in \cite{shorten2014evolvable}. It may not be high enough for the selection pressure towards evolvability to play its role, for instance if most individuals end up crashing against a wall near the starting position. For the biped locomotion experiment, the number of individuals is larger (500), but it may not be enough for this domain. In a run driven by the fitness and with an elitist selection algorithm, an efficient individual may survive long enough to generate efficient offspring. It will stay in the population as long as no higher performers appears. In NS, it may just not have enough time to generate enough of them for selection to properly work. Other divergent algorithms like MAP-Elites \cite{mouret2015illuminating} may not have this problem as one individual will stay in the archive as long as it is not replaced by a locally higher performer. 

\section{Conclusion}

Evolvability can be obtained for free in unbounded environments \cite{lehman2013evolvability}. We have defined evolvability as (1) the capacity to reach the largest part of the outcome space and (2) as uniformly as possible. Our experiments suggest that NS pushes towards these two dimensions of evolvability and \textit{in bounded outcome spaces}. It also maintains a high level of coverage and uniformity throughout the search. We have shown that this phenomenon results from the dynamics of the novelty criterion, pushing individuals to perpetual movements in the outcome space, as a result of the constant changes in the reference set used to compute novelty.

\textbf{Acknowledgements:} 
This work has benefited from a support of the Labex SMART (ANR-11-LABX-65), supported by French state funds managed by the ANR within the Investissements d'Avenir programme under reference ANR-11-IDEX-0004-02.

\bibliographystyle{ACM-Reference-Format}
\bibliography{biblio} 

\end{document}